\documentclass[sigconf]{acmart}

\usepackage{multirow}
\usepackage{balance}

\AtBeginDocument{%
  }

\copyrightyear{2023}
\acmYear{2023}
\acmDOI{10.1145/3581783.3612040}

\setcopyright{acmlicensed}\acmConference[MM '23]{Proceedings of the 31st
ACM International Conference on Multimedia}{October 29-November 3,
2023}{Ottawa, ON, Canada}
\acmBooktitle{Proceedings of the 31st ACM International Conference on
Multimedia (MM '23), October 29-November 3, 2023, Ottawa, ON, Canada}
\acmPrice{15.00}
\acmISBN{979-8-4007-0108-5/23/10}

\acmSubmissionID{1528}

\begin{document}

\title{StyleEDL: Style-Guided High-order Attention Network for Image Emotion Distribution Learning}

\author{Peiguang Jing}
\email{pgjing@tju.edu.cn}
\orcid{0000-0003-2648-7358}
\affiliation{%
  \institution{School of Electrical and Information Engineering, Tianjin University}
  \streetaddress{}
  \city{Tianjin}
  \state{}
  \country{China}
  \postcode{}
}

\author{Xianyi Liu}
\email{goog@tju.edu.cn}
\orcid{0000-0001-6284-9470}
\affiliation{%
  \institution{School of Electrical and Information Engineering, Tianjin University}
  \streetaddress{}
  \city{Tianjin}
  \state{}
  \country{China}
  \postcode{}
}

\author{Ji Wang}
\email{wangji@tju.edu.cn}
\affiliation{%
  \institution{School of Electrical and Information Engineering, Tianjin University}
  \streetaddress{}
  \city{Tianjin}
  \state{}
  \country{China}
  \postcode{}
}

\author{Yinwei Wei}
\email{weiyinwei@hotmail.com}
\affiliation{%
  \institution{Faculty of Information Technology, Monash University}
  \streetaddress{}
  \city{}
  \state{Victoria}
  \country{Australia}
  \postcode{}
}

\author{Liqiang Nie}
\email{nieliqiang@gmail.com}
\affiliation{%
  \institution{School of Computer Science and Technology, Harbin Institute of Technology}
  \streetaddress{}
  \city{Shenzhen}
  \state{}
  \country{China}
  \postcode{}
}

\author{Yuting Su}
\email{ytsu@tju.edu.cn}
\affiliation{%
  \institution{School of Electrical and Information Engineering, Tianjin University}
  \streetaddress{Weijin Road}
  \city{Tianjin}
  \state{}
  \country{China}
  \postcode{43017-6221}
}

\renewcommand{\shortauthors}{Peiguang Jing et al.}

\begin{abstract}
Emotion distribution learning has gained increasing attention with the tendency to express emotions through images. 
As for emotion ambiguity arising from humans' subjectivity, substantial previous methods generally focused on learning appropriate representations from the holistic or significant part of images. 
However, they rarely consider establishing connections with the stylistic information although it can lead to a better understanding of images. 
In this paper, we propose a style-guided high-order attention network for image emotion distribution learning termed StyleEDL, which interactively learns stylistic-aware representations of images by exploring the hierarchical stylistic information of visual contents. Specifically, we consider exploring the intra- and inter-layer correlations among GRAM-based stylistic representations, and meanwhile exploit an adversary-constrained high-order attention mechanism to capture potential interactions between subtle visual parts.  
In addition, we introduce a stylistic graph convolutional network to dynamically generate the content-dependent emotion representations to benefit the final emotion distribution learning. 
Extensive experiments conducted on several benchmark datasets demonstrate the effectiveness of our proposed StyleEDL compared to state-of-the-art methods.
The implementation is released at: https://github.com/liuxianyi/StyleEDL.
\end{abstract}

\ccsdesc[500]{Networks~Network design principles}
\ccsdesc[500]{Networks~Network algorithms}

\keywords{emotion distribution learning, style-guided, high-order, stylistic GCN}


\maketitle

\section{Introduction}
Image emotion analysis~\cite{zhao2021affective} has gained significant research attention owing to its facility in conveying emotions and views of people. Currently, image emotion analysis  has been applied in various scenarios, such as multimedia retrieval~\cite{10.1145/3343031.3350999,DBLP:journals/tip/ZhuLCLZ20, 10144360, qu2021dynamic, nie2022search}, 
social network analysis~\cite{serrat2017social,freeman2004development,wasserman1994social, jing2023category}, advertising recommendation~\cite{yang2013integrated,Holbrook1984,mitchell1986effect}.

In prior studies, the image emotion analysis tends to be formulated as a single-label classification task~\cite{rao2020learning,yang2018weakly,yang2018visual,zhang2019exploring,zhu2017dependency}, where each image is assigned a dominant label. 
However, one image may contain a mixture of multiple emotions with varying intensities, and an individual may have different emotional responses toward one image (i.e., ambiguity). 
As to this problem, the label distribution learning (LDL) paradigm ~\cite{yang2017joint,fan2018predicting,wang2021label,gao2017deep} has been adopted to narrow the gap between visual features and affective states.
Typically, ~\cite{yang2017joint} intended to learn a more smooth label vector to represent the emotions of images, replacing the previous dominant emotion classification.
~\cite{fan2018predicting} attempted to boost predicting performance by taking regions that represent emotions most into consideration.
However, these methods failed to explicitly consider the correlations between emotions. For example, an image of a reunion of old friends may be more likely to evoke feelings of both excitement and happiness, without causing sadness. 
Fortunately, emotion correlations~\cite{yang2021circular,he2019image,xiong2019structured,xu2021emotional} have been proven to be able to further improve the emotional distribution performance with prior knowledge. 
\begin{figure}[htbp]
    \centering
    \includegraphics[width=0.48\textwidth]{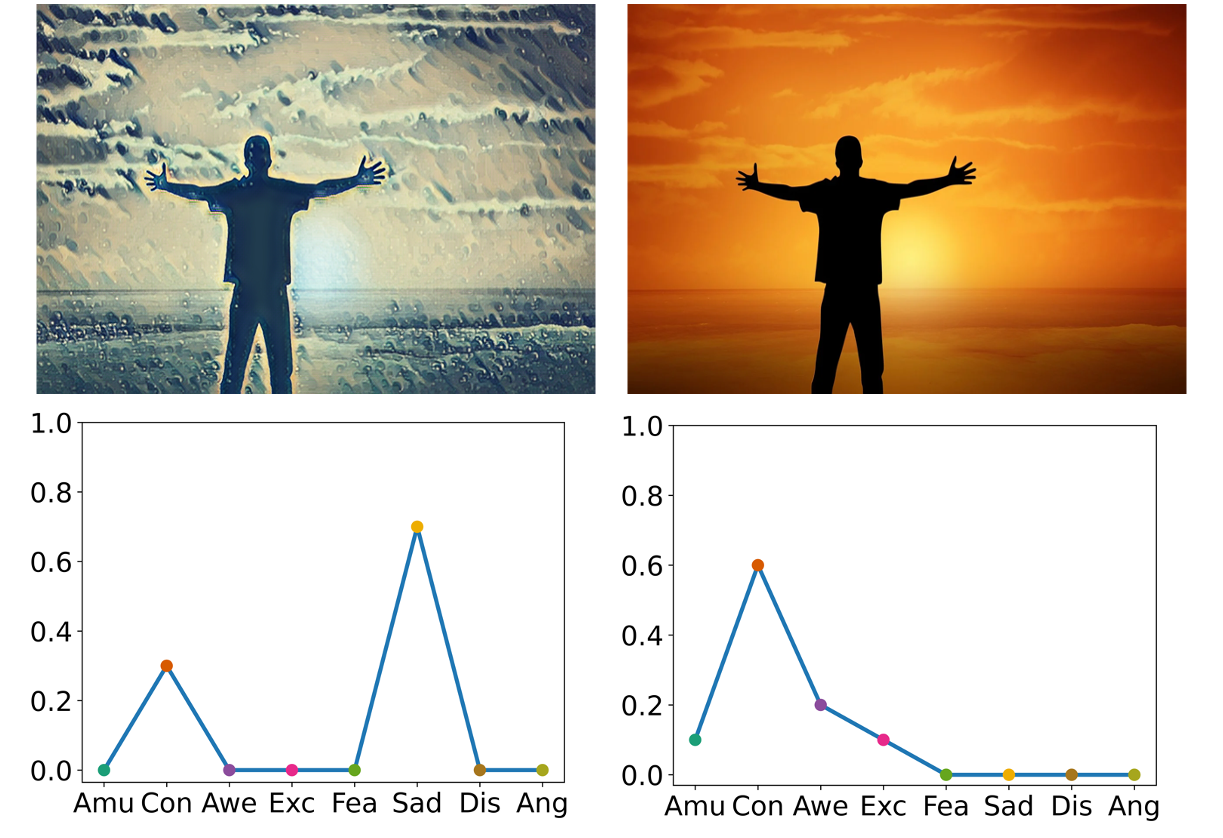}
    \caption{Different styles of images can elicit different emotional responses. In the case where the content is the same but the style is different, the left side contains more melancholic while the right side is more content.}
    \label{intro}
    \vspace{-0.5em}
\end{figure}

However, existing methods for emotion distribution learning usually suffer from two challenges that tightly hinder performance improvements.
First, due to the subjectivity of human cognition, directly using visual representations extracted from convolutional neural networks (CNNs) may be insufficient to characterize emotions contained in images, especially for the emotion ambiguity problem. 
As an example, Figure~\ref{intro} shows that the right side has more contentment, while the left side may evoke melancholic feelings among observers, despite depicting the same content. 
Different styles shown in these two pictures cause different emotions, but the existing methods rarely reveal this point from the perspective of stylistic representation learning.
Second, the correlation among emotions is generally modeled by a static graph structure~\cite{he2019image}.
Unfortunately, the adjacent relation of a static graph is usually manually defined according to the given dataset, and such relation generally models coarse dependencies, with limited versatility to mine fine latent relationships between emotions. 
As a result, these methods learn only coarse  emotional correlations during iterations, which ultimately leads to unsatisfactory prediction performance.

To address the above issues, we propose a novel method termed style-guided high-order attention network for image emotion distribution learning (StyleEDL).
The core idea behind our proposed StyleEDL is to leverage stylistic information to compensate for the deficiency of visual representations to resolve emotion ambiguity.
To explore stylistic-aware information in datasets, we first use GRAM-based intra- and inter-layer correlations as emotional style representations.
And then we intermix significant attention results of content information generated by an adversary-constrained high-order attention module to obtain stylistic-aware representations.
To get more accurate emotions of images, we consider the intrinsic relationship upon stylistic-aware representations using a stylistic graph network.
Taking the coarse information of static graph network learning as the prior information, the network adopts a dynamic graph structure to obtain the emotional-aware representations from the stylistic-aware representations in an adaptive way.
Our main contributions are as follows:
\begin{itemize}
    \item We propose a novel emotion distribution learning method termed StyleEDL, which explores stylistic information as complementary information to refine the representations of images.
    To the best of our knowledge, this is the first work using a style-induced paradigm for IEDL.
    \item We devise a stylistic-aware representation learning network that extracts attentive visual content representations and hierarchical stylistic representations.  In addition, develop a stylistic GCN to capture the intrinsic correlation among stylistic-aware representations.
    \item We conduct experiments on three public datasets, and the results show the superiority of the proposed method compared to state-of-the-art methods.
\end{itemize}
The remaining sections of the paper have been structured as follows:
Section 2 expounds on the related research,
Section 3 presents the proposed StyleEDL,
Section 4 provides empirical evaluation and analysis, and
Section 5 concludes the paper.

\section{Related Work}
\subsection{Image Emotion Distribution Learning}
Existing research on LDL can be borrowed to describe the emotions corresponding to an image. In particular, 
\cite{yang2017learning} proposed two methods, conditional probability neural network with binary code (BCPNN) and augmented conditional probability neural network (ACPNN), based on conditional probability neural networks to address sentiment ambiguity with multiple emotions. 
\cite{gao2017deep} proposed the deep label distribution learning (DLDL) method, which effectively utilizes label ambiguity by minimizing the Kullback-Leibler divergence for the first time. 
\cite{yang2017joint} developed a multi-task deep framework by jointly optimizing classification and distribution prediction. 
Later, polarity and relevance among emotions were also taken into account to explicitly model emotional correlation, making it effective to learn the distribution. 
\cite{he2019image} utilized graph neural networks as emotional predictors to capture the correlation among emotions.
\cite{xiong2019structured} designed a combined loss based on the earth mover's distance (EMD) and kullback-leibler divergence using structured labels in sentiment polarity. 
\cite{yang2021circular} designed a novel progressive circular (PC) loss based on an emotional circle to boost the learning process in an emotion-specific way.
To explore emotional style representations in complicated images, our proposed method proposes stylistic-aware representation learning and emotional-aware enhanced representation learning, producing accurate emotion distribution in real-world datasets. 
\subsection{Image Style Recognition}
Many recent works have indicated that the style of an image has a significant impact on the meaning it conveys. For example, 
\cite{lu2015deep} proposed a multi-patch aggregation network for extracting fine-grained features from images and showed that this approach achieves good performance in image style classification, aesthetic classification, and quality estimation tasks. 
\cite{matsuo2016cnn} used the GRAM matrix of feature maps to generate style vectors, which they applied to style image retrieval. 
\cite{lecoutre2017recognizing} demonstrated the effectiveness of deep residual networks in image style recognition. 
\cite{yang2018historical} proposed a multi-factor distribution soft label and performed image style classification in a multi-task framework. 
\cite{chu2018image} systematically explored the use of correlations between feature maps to characterize image style.
\cite{laubrock2019cnn} confirmed that mid-level features corresponding to textures, shadows, etc., are particularly well-suited for illustration style classification. 
\cite{ghosal2019geometry} proposed using geometry-sensitive style features based on image saliency for photographic image classification.
However, those works have only focused on directly extracting feature maps from their models, which may not fully capture fine representations. In this paper, we propose a novel style-induced method that leverages attentive visual content representations and hierarchical stylistic representations to guide emotion distribution learning. This approach allows for more comprehensive emotional representations than previous methods.

\begin{figure*}[!ht]
    \centering
    \includegraphics[width=1\textwidth]{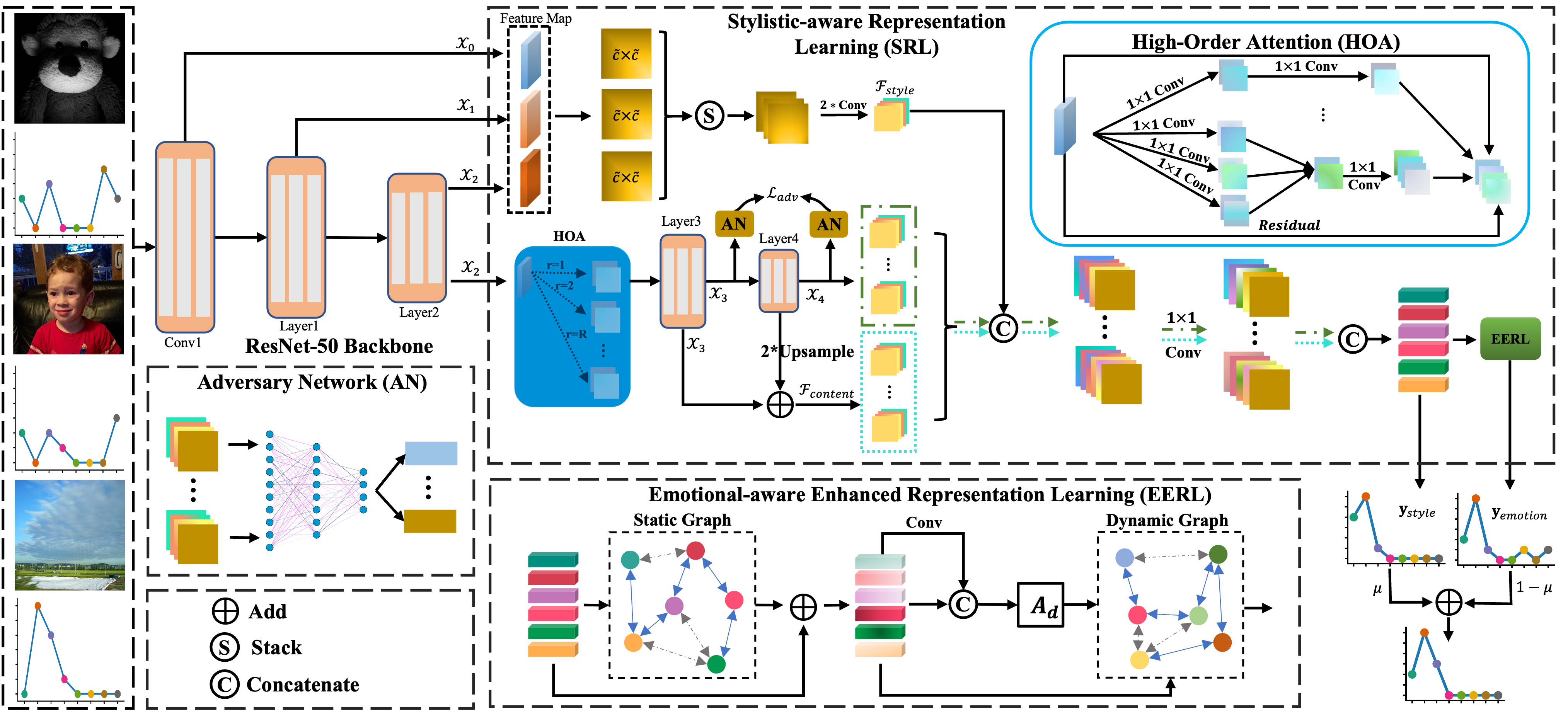}
    \caption{Detailed structure of our StyleEDL, which consists of three core networks: 
    (1) ResNet-50 is the backbone of our method, discarded the last fully connected layer and retained the top convolutional layer and four groups of convolutional layers, namely `Conv1', `Layer1', `Layer2', `Layer3', and `Layer4'.
    (2) Stylistic-aware representation learning network generates stylistic-aware representations based on emotional content representations~$\mathcal{F}_{content}$~and emotional style representations~$\mathcal{F}_{style}$.
    (3) Emotional-aware enhanced representation Learning network uses a stylistic GCN to obtain emotional-aware enhanced representations.
    }
    \label{framework}
    \vspace{-0.5em}
\end{figure*}

\section{The Proposed Method}
Emotion distribution learning task can be defined as: given a labeled sample pair $\{x,\hat{\mathbf{y}}\}$, which is used to learn a function:
\begin{equation}
    \small
    {\mathcal{H}: x \xrightarrow{} \mathbf{y} }
\end{equation}
where $x$ represents an input image, $\hat{\mathbf{y}} =\{\hat{y}_n\}_{n=1}^{N}$ ($\sum\nolimits_{n=1}^N \hat{y}_n^i= 1$, $\hat{y}_n^i \in [0,1]$) is the degree to which $N$ emotions are expressed in this image, and $\mathbf{y}$ represents the corresponding emotional distribution learned.
Our goal is to optimize the function $\mathcal{H}$ with the help of supervised information $\hat{\mathbf{y}}$, fitting the true emotional distribution of the image.
\subsection{Framework Overview}
The representations of images can evoke different emotions depending on the aspects being considered. 
Different aspects can also contribute differently to the triggering of emotions. 
One way to construct different representations was to directly use shallow features extracted by CNN as emotional concepts.
However, these concepts may not fully capture the emotional content of an image.
Another way to represent emotions in different aspects is by using CNN with multiple convolutional layers. 
Even though a convolutional layer with small kernels may struggle to perceive everything in an image, a deeper architecture can increase the model's receptive field. 
As a result, the early layers of the CNN tend to capture low-level features such as color and texture, while the later layers capture more complex and high-level features.
Therefore, we use the characteristics of the CNN to construct a module for the \textbf{stylistic-aware representation learning}.
In detail, we first use the GRAM matrix of low-level features as stylistic information.
And then we combine it with visual content information from high-level features enhanced by a high-order attention module to obtain stylistic-aware representations.
Moreover, recent studies~\cite{mittal2021eclare} have shown that the graph convolutional network (GCN) can improve the performance of emotion distribution learning due to its ability to capture emotion dependencies. 
However, traditional GCN only captures coarse emotion dependencies.
And the stylistic-aware representations contain relatively comprehensive emotional representations from visual and stylistic information, but only parts of them play a role in improving performance.
Therefore, we propose a stylistic GCN module for \textbf{emotional-aware enhanced representation learning}, which captures emotion relations of stylistic-aware representations in an adaptive way.
Specifically, the module initializes coarse emotion dependencies from a static GCN and uses them to capture emotional-aware dependencies of stylistic-aware representations for each image. 
By integrating the stylistic-aware representations and emotional-aware dependencies, our proposed method can better capture the emotions present in images.

\subsection{Stylistic-aware Representation Learning}
In this module, we first learn the emotional style representations with image style by exploring intra-layer and inter-layer correlations in feature maps. Second, a high-order attention mechanism with adversary constraints is introduced to guide learning emotional content representations. 
Finally, we fuse the emotional style representations and emotional content representations and further explore the latent stylistic-aware representations of images.
\subsubsection{GRAM-based Intra- and Inter-layer Correlation}
Inspired by the work~\cite{matsuo2016cnn}, we use the GRAM matrix of each layer's feature maps as the intra-layer emotional style representations. Because the features related to the emotional style of the image, such as texture and color, are typically captured in the low-dimensional feature maps, we first extract the input feature maps from the outputs of different layers of ResNet-50 to calculate the stylistic representation within each layer, as follows:
\begin{equation}
    \small
    {\mathcal{X}_k =[\mathbf{X}^1_k,\mathbf{X}^2_k,...,\mathbf{X}^{c_k}_k]}\in {\mathbb{R}^{c_k \times w_k \times h_k}}, k=0,1,2
\end{equation}
where $\mathcal{X}_k$ represents the feature maps of the $k$-th layer extracted from an image $x$.
For simplicity, in all the following layers, $c_*$, $w_*$ and $h_*$ represent the number of channels, width and height of the feature maps or representations, respectively.
We then convert feature maps into a vector $\mathbf{x}^{i}_k \in \mathbb{R}^{m_k}$,$i = 0,1,...,c_k$,$m_k = w_k \times h_k$
and concatenate them into a matrix $\mathbf{B}_k$.
\begin{equation}
    \small
    \mathbf{B}_k = [\mathbf{x}^1_k,\mathbf{x}^2_k,...,\mathbf{x}^{c_k}_k] \in \mathbb{R}^{c_k \times m_k}
\end{equation}
Following the above transformation, we can obtain the GRAM matrix of each layer as the corresponding intra-layer correlation.
\begin{equation}
    \small
    \mathbf{G}_k =\mathbf{B}_k {\mathbf{B}_k}^\mathrm{T} \in \mathbb{R}^{c_k \times c_k}, k=0,1,2
    \label{gram}
\end{equation}
In $\mathbf{G}_k$, each element $G^{ij}_k = \sum_a{\mathbf{B}^{ia}_k}{\mathbf{B}^{aj}_k}, a=0,1,\cdots,m_k$ is the inner product between the transformed feature maps $i$ and $j$ in layer $k$.

To capture the correlations between different layers of the network, we first use the GRAM matrix $\mathbf{G}^k$ defined in Eq.~(\ref{gram}) to obtain the emotional style representation for each layer. 
However, the shape of the GRAM matrix obtained from different layers may vary, so we next upsample them to the same shape and further stack them together along the channel dimension, denoted as
\begin{equation}
    \small
    \widetilde{\mathcal{G}} = \mathrm{Stack}( \widetilde{\mathbf{G}}_0, \widetilde{\mathbf{G}}_1, \widetilde{\mathbf{G}}_2)
\end{equation}
\begin{equation}
    \small
    \widetilde{\mathbf{G}}_k = \mathrm{Upsample}(\mathbf{G}_k), k=0,1,2
\end{equation}
where $\widetilde{\mathcal{G}}$ is the stacked GRAM matrix by the operator $\mathrm{Stack}(\cdot)$.

To measure the inter-layer correlations in the image, we design an inter-layer correlation module that consists of two convolutional layers, each followed by a layer normalization (LN) and a ReLU activation.
Unlike the commonly used instance normalization (IN) in image style transfer networks, the proposed module uses LN to normalize the input feature maps along the three dimensions of channel, width, and height. 
This increases the correlation between channels, and can be expressed as follows:
\begin{equation}
    \small
    \mathcal{F}_{style}  = f_{cc}( \widetilde{\mathcal{G}} ) \in {\mathbb{R}^{c_s \times w_s \times h_s}}
\end{equation}
where $\mathcal{F}_{style}$ is the emotional style representations, $f_{cc}$ represents inter-layer correlation module.

\subsubsection{Visual Attention Module}
The visual content information of an image can be derived from the objects and scene information depicted in the image.
Complex emotions are not easily acquired from those pieces of information.
Inspired by~\cite{chen2019mixed} in solving person re-identification tasks, we propose a visual attention module that introduces high-order attention (HOA) and feature pyramid mechanisms.
It captures the complex relations and subtle differences among visual parts to enhance the ability of emotion distribution learning. 
Specifically, 
given the feature maps $\mathcal{X}_{2}$ obtained from the `Layer2' layer, HOA is utilized to model the high-order attentive results $\mathcal{X}_{att}^r$ among visual parts as follows:
\begin{equation}
    \small
    \mathcal{X}_{att}^{r} = f_{hoa}^r(\mathcal{X}_{2}), r=1,\ldots, R
\end{equation}
\begin{equation}
    \small
   f_{hoa}^r(\mathcal{X}) = \sum_{s=1}^{r}\mathrm{Conv_{1\times1}}(\mathcal{Z}_1^r \odot \cdots \odot \mathcal{Z}_s^r \odot \cdots \odot \mathcal{Z}_r^r)
\end{equation}
\begin{equation}
    \small
    \label{hoa2}
   \mathcal{Z}_s^r = \mathrm{Conv_{1\times1}}(\mathcal{X}), s=1,\cdots,r
\end{equation}
where $f_{hoa}^r$ represents HOA, $R$ is the number of order, $\odot$ represents element-wise product operator. 
Specifically, Eq.~(\ref{hoa2}) mines simple and coarse information from $s$ emotional perspectives using various $1\times1$ convolution layers $\mathrm{Conv_{1\times1}}(\cdot)$. The element-wise product operator that follows is used to capture the complex and high-order interactions of visual parts, as well as the subtle differences among emotional-aware regions caused by the objects present in the image.
As shown in Figure~\ref{framework}, to generate the emotional content features under the guidance of high-order relationships, we use the `Layer3' layer termed $f_{3}(\cdot)$ and `Layer4' layer termed $f_{4}(\cdot)$ in ResNet-50 to encode the high-order attentive results to obtain multi-scale emotional content features $\mathcal{X}_{3}$ and $\mathcal{X}_{4}$:
\begin{equation}
    \small
    \mathcal{X}_{3} = [f_{3}(\mathcal{X}_{att}^{1}),\ldots ,f_{3}(\mathcal{X}_{att}^{R})] \in \mathbb{R}^{R \times c_{3} \times w_{3} \times h_{3} }
\end{equation}
\begin{equation}
    \small
    \mathcal{X}_{4} = [f_{4}(\mathcal{X}_{3}^{1}),\ldots, f_{4}(\mathcal{X}_{3}^{R})] \in \mathbb{R}^{R \times c_{4} \times w_{4} \times h_{4} }
\end{equation}
To effectively extract and describe visual content information, we construct a feature pyramid network (FPN) to improve the network's multi-scale perception ability. This is done by upsampling the feature maps $\mathcal{X}_{4}$ and convolving it with a $1\times1$ convolution layer to match the number of channels in $\mathcal{X}_{3}$. The resulting feature maps are then added to $\mathcal{X}_{3}$ to obtain the emotional content representations $\mathcal{F}_{content}$.
\begin{equation}
    \small
    \mathcal{F}_{content} = \mathrm{Conv_{1\times1}}(\mathrm{Upsample}(\mathcal{X}_{4}))+\mathcal{X}_{3}
\end{equation}
The HOA can explicitly capture diverse and complementary high-order information, which encourages the richness of the learned features.
However, simply employing the HOA module causes partial/biased learning behavior, hindering the performance of our method. The variant labeled as "noAN" aptly demonstrates this fact with great efficacy in Table~\ref{ablation}.   
As mentioned in~\cite{chen2019mixed}, we introduce an adversary constraint to suppress the problem of order collapse for the multi-scale emotional content features $\mathcal{X}_{3}$ and $\mathcal{X}_{4}$, respectively:
\begin{equation}
    \small
\begin{aligned}
    &\max\limits_{HOA\vert^{R=r}_{R=1}} \min\limits_{f_{adv}} \mathcal{L}_{adv}^{k} =  \\
    &\max\limits_{HOA\vert^{R=r}_{R=1}} \min\limits_{f_{adv}} (\sum_{s,s^{\prime}=1,s\neq s^{\prime}}^r \parallel f_{adv}(\mathbf{x}_k^s) - f_{adv}({\mathbf{x}_k^{s^\prime}}) \parallel^2_2)) 
    \label{adv2}
\end{aligned}
\end{equation}
\begin{equation}
    \small
    \mathcal{X}_k=[\mathcal{X}_k^1, \dots, \mathcal{X}_k^r, \dots, \mathcal{X}_k^R], k=3,4
\end{equation}
\begin{equation}
    \small
    \mathbf{x}_k^r = \mathrm{Flatten}(\mathcal{X}_{k}^r)
\end{equation}
where $f_{adv}$ is the adversary network (AN) which contains two fully-connected layers, $HOA\vert^{R=r}_{R=1}$ means there are $r$ HOA modules (from first-order to $r$-th order), $\mathbf{x}_k^s\in\mathbb{R}^{D_s} (D_s = c_k\times w_k \times h_k)$ is the multi-scale emotional content vector flattened from $\mathcal{X}_k^r$ with $r=s$ and $\mathrm{Flatten}(\cdot)$ is the flattening operator.
According to Eq.~(\ref{adv2}) , we get the adversarial loss $\mathcal{L}_{adv}=\sum\nolimits_k \mathcal{L}_{adv}^{k}$.

\subsubsection{Stylistic-aware Representation}
Based on the emotional style and content representations learned above, we use a fusion operator to obtain the stylistic-aware distribution $\mathbf{y}_{style}$. The fusion operator is a $1\times1$ convolution layer with concatenation, which produces an output with the same number of channels as the number of emotion categories.

Specifically, we first use the concatenation operator to combine the stylistic-content representation pairs $\{\mathcal{F}_{style}, \mathcal{F}_{content}\}$ and $\{\mathcal{F}_{style}, \mathcal{X}_{4}\}$ to obtain the intermediate representations $\mathcal{F}_{sc} \in \mathbb{R}^{R \times c_{sc} \times w_{sc} \times h_{sc}}$ and $\mathcal{F}_{s4} \in \mathbb{R}^{R \times c_{s4} \times w_{s4} \times h_{s4}}$, respectively. 
On the condition, we set $C=c_{sc}=c_{s4}$.
We then concatenate these intermediate representations to obtain the stylistic-aware representations $\mathcal{F}_e \in \mathbb{R}^{R \times C \times D_e}$, where $D_e = h_{sc} \times w_{sc} + h_{s4} \times w_{s4}$.
Finally, we apply global average pooling $\mathrm{mean}(\cdot)$ and global max pooling $\mathrm{max}(\cdot)$ to $\mathcal{F}_e$ to generate the stylistic-aware distribution results $\mathbf{y}_{style}$.
\begin{equation}
    \small
    \label{out0}
    [\mathbf{y}_{e}^1, \cdots, \mathbf{y}_{e}^R] = \mathrm{Softmax}(\mathrm{mean}(\mathcal{F}_e) + \lambda \ast \max(\mathcal{F}_e)) \in \mathbb{R}^{ R \times C }
\end{equation}
\begin{equation}
    \small
    \mathbf{y}_{style} = \mathrm{mean}(\mathbf{y}_{e}^1, \cdots, \mathbf{y}_{e}^R)
    \label{out1}
\end{equation}
where $\lambda$ is the coefficient to control the trade-off between two types of pooling method, $\mathrm{Softmax}(\cdot)$ is the activation function to unify the element value in $\mathbf{y}_{style}$ to $[0,1]$.
\subsection{Emotional-aware Enhanced Representation Learning}
Different from other LDL tasks, emotions and their unique characteristics have intrinsic relationships, as demonstrated in psychological theories~\cite{yang2021circular}. Previous work~\cite{yang2021circular,he2019image,xiong2019structured} has shown that exploiting the correlations between emotion labels can improve the prediction of the emotion distribution of images.

Inspired by~\cite{ye2020attention}, we introduce a stylistic GCN, which consists of a static GCN termed $f_{SGCN}$ and a dynamic GCN termed $f_{DGCN}$, obtaining initialization representation~$\mathbf{F}_{sgcn}$ and emotional-aware enhanced representations~$\mathbf{F}_{dgcn}$ as follows:
\begin{equation}
    \small
    \mathbf{F}_{sgcn} = f_{SGCN}(\mathbf{A}_s, \widetilde{\mathbf{F}}_e, \mathbf{W}_s)
\end{equation}
where $\mathbf{A}_s$ is the graph adjacency matrix constructed using the co-occurrence relationship between labels. $\widetilde{\mathbf{F}}_{e} \in \mathbb{R}^{C \times D}$ is obtained by concatenating $\mathcal{F}_{e}$, where $D = D_e^1 + D_e^2 + \ldots+ D_e^r + \ldots + D_e^R$. $D_e^r$ represents $r$-th order stylistic-aware representations. $\mathbf{W}_s$ are learnable parameters.
However, the static GCN is not very flexible and can not eliminate irrelevant information of stylistic-aware representations to capture fine emotional dependencies.
Therefore, we use the adaptability of the dynamic graph network to better capture emotional-aware enhanced representations:
\begin{equation}
    \small
    \mathbf{F}_{dgcn} = f_{DGCN}(\mathbf{A}_d, \mathbf{F}_{sgcn}, \mathbf{W}_d)
\end{equation}
\begin{equation}
    \small
    \mathbf{A}_d = \delta (\mathbf{W}_A \widetilde{\mathbf{F}}_{dgcn})
\end{equation}
where $\mathbf{A}_d$ enables the network structure to be dynamically adjusted for each image. $\mathbf{W}_A$ and $\mathbf{W}_d$ are learnable parameters. $\widetilde{\mathbf{F}}_{dgcn}$ is obtained by concatenating $\mathbf{F}_{sgcn}$ and its global representations $\mathbf{F}_{sgcn}$, $\delta(\cdot)$ is the sigmoid activation function.

In the same way as Eq.~(\ref{out0}) and Eq.~(\ref{out1}), we can obtain the emotional-aware distribution results $\mathbf{y}_{emotion} \in \mathbb{R}^{C}$ of the module.

\subsection{Final Distribution and Optimization}
Once we have these two predicted distributions $\mathbf{y}_{emotion}$ and $\mathbf{y}_{style}$, we can simply combine them using the weighted sum defined above to obtain the final emotional distribution $\mathbf{y}$ as follows:
\begin{equation}
    \small
    \mathbf{y} = \mu \ast \mathbf{y}_{emotion} + ( 1 - \mu ) \ast \mathbf{y}_{style}
    \label{fo}
\end{equation}
where $\mu$ is the coefficient to control the trade-off between two different predicted results.

As with the previous method, the proposed method employs the KL loss~\cite{gao2017deep} for the distribution learning.
Our objective function consists of adversarial loss $\mathcal{L}_{adv}$ and prediction loss $\mathcal{L}_{pred}$.
For prediction loss, we consider intermediate supervision instead of directly optimizing the predicted results as follows:
\begin{equation}
    \small
    \mathcal{L}_{pred} =\mathrm{mean} (\frac{1}{R}  \sum_{r=1}^R (\mathrm{KLloss}(\mathbf{y}_e^r, \hat{\mathbf{y}})) + \mathrm{KLloss}(\mathbf{y}_{emotion}, \hat{\mathbf{y}}))
\end{equation}
Meanwhile, in order to balance the difference in the numerical scale of the two losses, we adopt an adaptive balance method:
\begin{equation}
    \small
    \mathcal{L} = \mathcal{L}_{pred} +  \mathcal{L}_{adv} / \parallel \mathcal{L}_{adv}/\mathcal{L}_{pred}\parallel
\end{equation}
where $\parallel \mathcal{L}_{adv}/\mathcal{L}_{pred}\parallel$ represents the truncated gradient operator, which calculates the adaptive balance coefficient of adversarial loss.

\section{EXPERIMENT}
\subsection{Experimental Setup}
\subsubsection{Emotion Datasets}
\textbf{Flickr-LDL}~\cite{yang2017learning} is a collection of images that has been annotated with emotional label distributions (\textit{i.e.anger, amusement, awe, contentment, disgust, excitement, fear and sadness}). 
Per emotional category was created by selecting a subset of the Flickr dataset~\cite{borth2013large} using 1200 adjective-noun pairs, and then having 11 viewers annotate the images with one of eight common emotions.
The final dataset contains 10700 images, with roughly equal numbers of images per emotion class.
\textbf{Twitter-LDL}~\cite{yang2017learning} was created by 
using a variety of emotional keywords to search for images on Twitter and then the retrieved images were manually screened and annotated by 8 viewers.
The final dataset contains 10045 images, with the annotations indicating the distribution of emotions present in each image.
\textbf{Emotion6}~\cite{peng2015mixed} contains 1980 images that were obtained from Flickr using seven categories of emotion keywords (\textit{i.e.anger, disgust, joy, fear, sadness, surprise and neutral}), with 330 images in each category. And each image was annotated by 15 viewers.

\begin{table*}
    \vspace{-1em}
    \small
    \begin{center}
        \caption{Comparison with the state-of-the-art methods on Twitter-LDL dataset.}
        \label{result1}
            \begin{tabular}{ccccccccccc}
                \toprule
                \toprule   
                                           Measures
                                           & PT-Bayes                 & PT-SVM
                                           & AA-kNN                  & AA-BP
                                           & SA-BFGS                  & SA-CPNN
                                           & SSDL                    & LDL-LDM        & DIEDL         
                                           & Ours    \\

                \midrule 
                                           KL $\downarrow$         
                                           & 1.31(8)   & 1.65(9)
                                           & 3.89(10)             & 1.19(6)
                                           & 1.19(6)              & 0.85(5)
                                    
                                           & 0.51(3)               & 0.53(4)  & 0.47(2)               & \textbf{0.42(1)} \\

                                           Chebyshev $\downarrow$  
                                           & 0.53(9)   & 0.63(10)
                                           & 0.28(5) 
                                           & 0.37(7)
                                           & 0.37(7)      & 0.36(6)
                                           & 0.25(3)
                                           &0.27(4) 
                                           & 0.24(2)
                                           & \textbf{0.22(1)} \\

                                           Clark $\downarrow$ & 0.85(5)
                                           & 0.91(9)
                                           & \textbf{0.58(1)}
                                           & 0.89(7)
                                           & 0.89(7)
                                           & 0.85(5)
                                        
                                           & 0.84(2)
                                           & 2.35(10)
                                           & 0.84(2)
                                           & 0.84(2) \\

                                           Canberra $\downarrow$ 
                                           & 0.77(3)
                                           & 0.88(9)
                                           & \textbf{0.41(1)}
                                           & 0.84(7)
                                           & 0.84(7)
                                           & 0.78(6)
                                           & 0.76(2)
                                          
                                           & 6.05(10)
                                           & 0.77(3)
                                           & 0.77(3) \\

                                           Cosine $\uparrow$       & 0.53(9)                     & 0.25(10)
                                           & 0.82(5)                 & 0.71(8)
                                           & 0.82(5)                    & 0.75(7)
                                           
                                           & 0.86(3)                    & 0.85(4) 
                                           & 0.87(2)
                                           & \textbf{0.89(1)} \\

                                           Intersection $\uparrow$ & 0.40(9)   & 0.21(10)
                                           & 0.66(5)              & 0.59(6)
                                           & 0.57(7)              & 0.56(8)
                                          
                                           & 0.69(2)                     & 0.67(3)  
                                           & 0.67(4)
                                           & \textbf{0.73(1)} \\

                                           Average Rank $\downarrow$ & 7.17(9) & 9.50(10) & 4.50(4) & 6.83(7) & 6.50(6) & 6.17(8) & 2.50(2) & 5.83(5) & 2.50(2) &\textbf{1.50(1)} \\

                \bottomrule
                \bottomrule
            \end{tabular}
    \end{center}
    \vspace{-1em}
\end{table*}
\begin{table*}
    \small
    \begin{center}
        \caption{Comparison with the state-of-the-art methods on Emotion6 dataset.}
        \label{result2}
            \begin{tabular}{ccccccccccc}
                \toprule
                \toprule
                            
                                           Measures
                                           & PT-Bayes                 & PT-SVM
                                           & AA-kNN                  & AA-BP
                                           & SA-BFGS                  & SA-CPNN
                                
                                           & SSDL                     & LDL-LDM        & DIEDL        
                                           & Ours          \\

                \midrule  
                                           KL $\downarrow$   
                                           & 2.32(10)                     
                                           & 1.07(8)
                                           & 0.85(7)                    & 0.63(6)
                                           & 1.16(9)                     & 0.56(5)
                                          
                                           & 0.40(2)                    & 0.44(4)
                                           & 0.40(2)
                                           & \textbf{0.36(1)} \\

                                           Chebyshev $\downarrow$  & 0.35(8)                     & 0.39(10)
                                           & 0.29(5)                    & 0.30(6)
                                           & 0.38(9)                     & 0.30(6)
                                           
                                           & 0.24(2)                     & 0.26(3)    
                                           & 0.26(3)
                                           & \textbf{0.22(1)} \\

                                           Clark $\downarrow$
                                           & 0.73(8)
                                           & 0.69(7)
                                           & 0.62(2)
                                           & 0.64(6)
                                           & 0.74(9)
                                           & 0.63(5)
                                          
                                           & 0.62(2)
                                           & 1.65(10)
                                           & 0.62(2)
                                           & \textbf{0.59(1)} \\ 
                                    
                                           Canberra $\downarrow$
                                           & 0.66(8)
                                           & 0.62(7)
                                           & 0.51(2)
                                           & 0.54(5)
                                           & 0.67(9)
                                           & 0.54(5)
                                          
                                           & 0.51(2)
                                           & 3.64(10)
                                           & 0.52(4)
                                           & \textbf{0.47(1)} \\

                                           Cosine $\uparrow$       & 0.69(6)                     & 0.48(10)
                                           & 0.75(4)                   & 0.68(7)
                                           & 0.63(9)                     & 0.66(8)
                                        
                                           & 0.79(3)                    & 0.72(5)
                                           & 0.81(2)
                                           & \textbf{0.84(1)} \\

                                           Intersection $\uparrow$ & 0.56(8)                     & 0.42(10)
                                           & 0.62(5)                    & 0.59(7)
                                           & 0.52(9)                     & 0.60(6)              
                                        
                                           & 0.66(2)                     & 0.65(4)
                                           & 0.66(2)
                                           & \textbf{0.70(1)} \\

                                           Average Rank $\downarrow$ & 8.00(8) & 8.67(9) & 4.17(4) & 4.17(4) & 9.00(10) & 5.83(6) & 2.17(2) & 6.00(7) & 2.50(3) &\textbf{1.00(1)} \\
                                           
                \bottomrule
                \bottomrule
            \end{tabular}
    \end{center}
    \vspace{-1em}
\end{table*}

\begin{table*}
    \small
    \begin{center}
        \caption{Comparison with the state-of-the-art methods on Flickr-LDL dataset.}
        \label{result3}
            \begin{tabular}{ccccccccccc}
                \toprule
                \toprule
                                    
                                           Measures
                                           & PT-Bayes                 & PT-SVM
                                           & AA-kNN                  & AA-BP
                                           & SA-BFGS                  & SA-CPNN
                                        
                                           & SSDL                     & LDL-LDM        & DIEDL         
                                           & Ours          \\

                \midrule
                   
                                           KL $\downarrow$          & 1.88(9)               
                                           & 1.69(8)
                                           & 3.28(10)                    & 0.82(5)
                                           & 1.06(6)                     & 1.06(6)
                                          
                                           & 0.46(3)                     & 0.49(2)
                                           & 0.46(3)
                                           & \textbf{0.39(1)} \\

                                           Chebyshev $\downarrow$  & 0.44(9)                     & 0.55(10)
                                           & 0.28(5)                    & 0.36(7)
                                           & 0.37(8)                     & 0.30(6)
                                          
                                           & 0.23(2)                     & 0.25(4)
                                           & 0.23(2)
                                           & \textbf{0.21(1)} \\

                                           Clark $\downarrow$
                                           & 0.89(9)
                                           & 0.87(8)
                                           & \textbf{0.57(1)}
                                           & 0.82(5)
                                           & 0.86(7)
                                           & 0.82(5)
                                          
                                           & 0.78(3)
                                           & 2.14(10)
                                           & 0.79(4)
                                           & 0.76(2) \\

                                           Canberra $\downarrow$
                                           & 0.85(9)
                                           & 0.83(8)
                                           & \textbf{0.41(1)}
                                           & 0.75(6)
                                           & 0.82(7)
                                           & 0.74(5)
                                          
                                           & 0.69(3)
                                           & 5.26(10) 
                                           & 0.70(4)
                                           & 0.66(2) \\

                                           Cosine $\uparrow$       & 0.63(9)                     & 0.32(10)
                                           & 0.79 (5)                   & 0.72(6)
                                           & 0.70(7)                     & 0.70(7)
                                          
                                           & 0.85(3)                  & 0.84(4)  
                                           & 0.86(2)
                                           & \textbf{0.88(1)} \\

                                           Intersection $\uparrow$ & 0.49(9)                     & 0.29(10)
                                           & 0.64(5)                    & 0.53(8)
                                           & 0.56(7)                     & 0.60(6)
                                          
                                           & 0.68(3)                     & 0.66(4)           
                                           & 0.70(2) & 
                                        \textbf{0.71(1)} \\

                                        Average Rank $\downarrow$ & 9.00(9) & 9.00(9) & 4.50(4) & 6.17(7) & 7.00(8) & 5.83(6) & 2.83(2) & 5.66(5) & 2.83(2) &\textbf{1.33(1)} \\
                \bottomrule
                \bottomrule
            \end{tabular}
    \end{center}
    \vspace{-1em}
\end{table*}

\subsubsection{Evaluation Metrics}
To evaluate the effectiveness of our proposed StyleEDL, four metrics are selected: Kullback-Leibler (KL) divergence, Chebyshev distance, Cosine coefficient, Intersection similarity, Clark distance and Canberra metric. Additionally, Average Rank is also adopted to indicate the total performance of each model.

\subsubsection{Parameter and Evaluation Settings}
We used a ResNet-50 model pre-trained on the ImageNet dataset as our backbone network and removed the last fully connected layer. 
We considered the outputs of the top convolutional layer and four groups of convolutional layers (`Conv1', `Layer1', `Layer2', `Layer3', and `Layer4') of the ResNet-50 model. 
All training images are resized to $448 \times 448$ pixels and undergo random scaling and horizontal flipping for data augmentation.
Our proposed method is implemented using the PyTorch deep learning framework and is trained on an NVIDIA GTX 1080Ti GPU. We used mini-batch stochastic gradient descent (SGD) with momentum and weight decay to optimize our proposed method. 
The mini-batch size is set to 8 and the learning rate is set to 0.01 for the first 10 epochs, then decreased 10-fold every 20 epochs until the total number of training epochs reaches 90.
\subsection{Experimental Results}
To evaluate the effectiveness of our proposed StyleEDL, we compared our proposed scheme with several existing state-of-the-art methods, which are grouped into four categories: 
problem transformation~(PT-Bayes and PT-SVM ~\cite{geng2016label}), 
algorithm adaptation~(AA-kNN and AA-BP~\cite{geng2016label}),
specialized algorithm~(SA-BFGS~\cite{geng2016label} and SA-CPNN~\cite{geng2013facial}) and
CNN-based methods~(SSDL~\cite{xiong2019structured}, LDL-LDM ~\cite{article} and DIEDL~\cite{10.1016/j.knosys.2022.110107}).
Table~\ref{result1},~\ref{result2} and \ref{result3} show the performances of these methods on three widely used datasets. The best results are highlighted in boldface. The down arrow $\downarrow$ next to the measure means a lower score is better, and the up arrow $\uparrow$ means that a higher one is better.
From the table, we can make the following observations: 
1) AA-kNN achieves insurmountable results in Clerk Distance and Canberra metric, affirming its superiority in addressing intersecting samples in visual emotion distributions.
2) CNN-based methods perform better than the other three types of algorithms, which suggests that CNNs have a stronger ability to capture emotional-related content information from visual parts. 
3) Our method consistently outperforms the other methods by a clear margin, indicating that we can expect more accurate results by considering stylistic representations in emotion distribution learning tasks. 
\subsection{Ablation Study}
To further investigate the influence of different components of the proposed method, several variants of our proposed method are configured and ablation experiments are conducted on the Twitter-LDL dataset for comparison. The variants of the model include:
(a) B only, which adopts the backbone network.
(b) B+G, which adds the GRAM-based intra- and inter-layer correlation based on the (a) model.
(c) B+V, which employs the visual attention module based on the (a) model.
(e) B+E, which adopts only backbone and emotional-aware enhanced representation learning. 
(d) B+G+V, which involves both GRAM-based intra- and inter-layer correlation and visual attention module based on the (a) model.
(f) B+G+V+E$^\star$, which replaces our emotional-aware enhanced representation learning  with a static GCN.
(g) B+G$^\star$+V+E, which only considers the correlation between GRAM matrices, but not the correlation within the layers.
(h) noAN, which discards adversarial constraint loss based on our proposed method. 
\begin{table}[tb]
    \vspace{-1em}
    \small
    \begin{center}
        \caption{Ablation analysis on Twitter-LDL. 
        `B', `G', `G$^\star$', `V', `E' and `E$^\star$' correspond to the backbone, GRAM-based intra- and inter-layer correlation, only GRAM-based inter-layer correlation, visual attention module, emotional-aware enhanced representation learning, and static GCN, respectively.}
        \label{ablation}
        \begin{tabular}{ccccc}
            \toprule
            \toprule
            Method  & KL$\downarrow$ & Chebyshev$\downarrow$ & Cosine$\uparrow$ & Intersection$\uparrow$ \\
            \midrule
            B       & 0.457          & 0.228           & 0.881            & 0.718                  \\
            B+G     & 0.448          & 0.228             & 0.881            & 0.719                  \\
            B+V     & 0.441          & 0.226          & 0.882            & 0.717                  \\
            B+E       & 0.482          & 0.229         & 0.876            & 0.718 \\ 
            B+G+V   & 0.433          & 0.223          & 0.884            & 0.719                  \\
            B+G+V+E$^\star$ & 0.465 & 0.228    & 0.879   & 0.715        \\
            B+G$^\star$+V+E & 0.434 & 0.224   & 0.884   & 0.719         \\
            noAN & 0.446 & 0.223   & 0.882   & 0.723         \\
            Ours & \textbf{0.420} & \textbf{0.218}    & \textbf{0.889}   & \textbf{0.726}         \\
            \bottomrule
            \bottomrule
        \end{tabular}
    \end{center}
    \vspace{-1.5em}
\end{table}
The results are shown in Table~\ref{ablation}.
In this table, we selected six metrics mentioned in our paper to report the emotion distribution based on their predicted results.
From the results, the following observations can be made:
1) Without style-induced information, B+V performs worse than B+G+V, indicating that the emotional style representations are beneficial for learning stylistic-aware representations.
2) B+G and B+V all take positive effects, which demonstrate that not only the learning of visual content information improves emotional distribution results, but also the style information plays an important role.
3) B+E yields inferior outcomes compared to B, suggesting that features extracted from ResNet-50 and directly applied to our stylistic GCN do not yield favorable results.
4) Our proposed StyleEDL consistently surpasses B+G+V, B+G+V+E$^\star$ and B+G$^\star$+V+E, which means our proposed method gains from the use of the intra- and inter-layer correlation and the stylistic GCN. Moreover, the outcomes further indicate that the flexible dynamic GCN can eliminate irrelevant information of stylistic-aware representations.
Similar observations also can be found for the other two datasets.
\begin{figure}[!t]
    \centering
    \includegraphics[width=8.5cm]{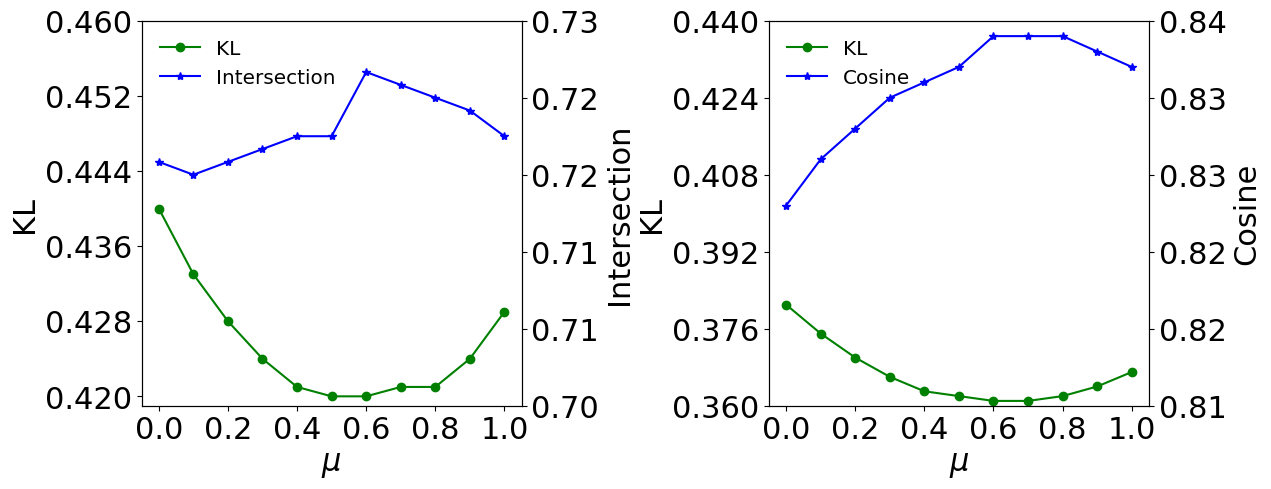}
    \caption{Effect of $\mu$ on datasets Twitter-LDL (left) and Emotion6 (right) }
    \label{mu}
\end{figure}
\begin{figure}[!t]
    \centering
    \includegraphics[width=8.5cm]{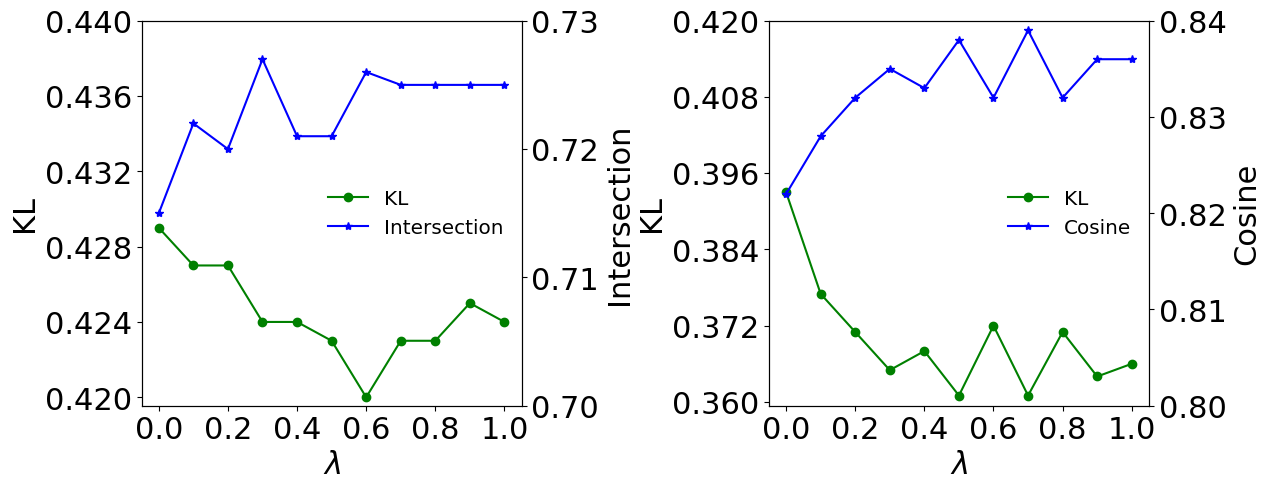}
    \caption{Sensitivity analysis of $\lambda$ on Twitter-LDL (left) and Emotion6 (right)}
    \label{lamda}
    \vspace{-1em}
\end{figure}
\subsection{Parameter Sensitivity Analysis}
In our work, there are three essential parameters, which are the order $R$ of the HOA module and the balance coefficients $\lambda$ and $\mu$ in stylistic-aware representation learning and emotional-aware enhanced representation learning, respectively.
We conducted comprehensive experiments on two datasets: the Twitter-LDL and the Emotion6. 
Specifically, the KL divergence and Intersection coefficient metrics  were used for the Twitter-LDL, while the KL divergence and Cosine coefficient metrics were used for the Emotion6.
\begin{table}[hb]
    \vspace{-1em}
    \small
    \begin{center}
        \caption{Sensitivity analysis of $R$ on Twitter-LDL.}
        \label{order_of_HOA_Emotion1}
        \begin{tabular}{ccccc}
            \toprule
            \toprule
            Order & $R=1$ & $R=2$ & $R=3$ & $R=4$\\
            \midrule
            KL$\downarrow$ & 0.445 & \textbf{0.420} & 0.421 &0.427 \\
            Chebyshev$\downarrow$ & 0.225 & \textbf{0.218} & 0.221 & 0.219\\
            Cosine$\uparrow$ & 0.882  & \textbf{0.889} & 0.888 & 0.886\\
            Intersection$\uparrow$ & 0.721  & \textbf{0.726}  & 0.720 & 0.724\\
            \bottomrule
            \bottomrule
        \end{tabular}
    \end{center}
    \vspace{-1em}
\end{table}
\begin{table}[hb]
    \small
    \begin{center}
        \caption{Sensitivity analysis of $R$ on Emotion6.}
        \label{order_of_HOA_Emotion2}
        \begin{tabular}{ccccc}
            \toprule
            \toprule
            Order & $R=1$ & $R=2$ & $R=3$ & $R=4$\\
            KL$\downarrow$ & 0.377 & \textbf{0.361} & 0.385 & 0.393\\   
            Chebyshev$\downarrow$  & 0.227 & \textbf{0.222}  & 0.231 & 0.235 \\
            Cosine$\uparrow$ & 0.829 & \textbf{0.839} & 0.827 & 0.822\\
            Intersection$\uparrow$  & 0.694 & \textbf{0.698}  & 0.689  & 0.687\\  
                   
            \bottomrule
            \bottomrule
        \end{tabular}
    \end{center}
    \vspace{-1em}
\end{table}

\subsubsection{Order of HOA module}
Tables~\ref{order_of_HOA_Emotion1} and ~\ref{order_of_HOA_Emotion2} show that our proposed method performs best when $R=2$ for both two datasets. 
When $R=1$, our proposed method lacks the ability to further encode feature maps and fails to employ the attention mechanism to refine the visual content representations, while a large value of $R$ makes the model more susceptible to being influenced by noise.

\begin{figure*}[!t]
    \centering
    \includegraphics[width=16cm, height=11.5cm]{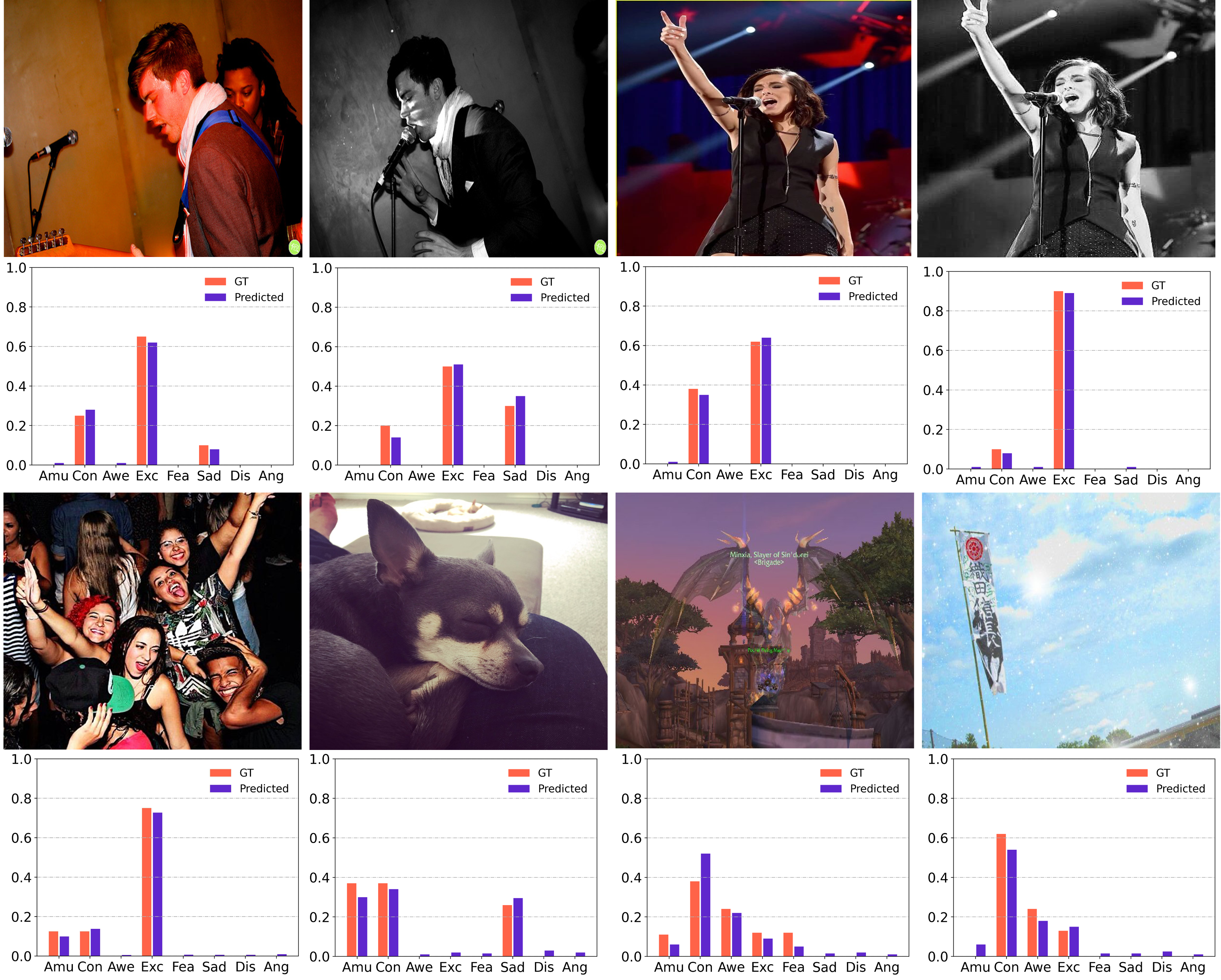}
    \caption{Visualization of the predicted emotion distributions (Predicted) and the ground-truth (GT). "Amu", "Con", "Awe", "Exc", "Fea", "Sad", "Dis" and "Ang" represent "amusement", "contentment", "awe", "excitement", "fear", "sadness", "disgust" and "anger" in Twitter-LDL, respectively.}
    \label{visualization}
    \vspace{-1em}
\end{figure*}
\subsubsection{Balance coefficient}
We investigated the influence of the balance coefficients $\lambda$ and $\mu$ by varying their values from $0.0$ to $1.0$. 
As shown in Figure~\ref{lamda}, larger values of $\lambda$ generally result in better performance than smaller values. 
A proper value of $\lambda$ can enhance the stylistic-aware representations of images and improve the overall performance of the model.
The coefficient $\mu$ plays an important role in balancing the importance between stylistic-aware distribution results and emotional-aware distribution results.
From Figure~\ref{mu},
our proposed method steadily improves from $0.0$ to $0.6$ and reaches its best performance at $0.6$. 
Intuitively, the performance can be enhanced by introducing emotional-aware enhanced representation learning.
Moreover, all values of KL divergence on emotion6 are much lower than that on Twitter-LDL, which may be owing to the fact that the dataset size of emotion6 is much smaller than that of Twitter-LDL.
\subsection{Computational Complexity}
Table~\ref{complexity} reports the actual inference time  with several recent state-of-the-art methods.
As discerned from the table, our approach achieves superior performance than those methods at the cost of the high computational complexity of the HOA, which proffers us a glimpse into the future.
In the future, we will explore light high-order solutions.
\begin{table}[b]
    \small
    \begin{center}
        \caption{Model complexity for inference with several state-of-the-art methods.}
        \label{complexity}
        \begin{tabular}{ccccc}
            \toprule
                & SSDL & LDL-LDM  &  DIEDL & Ours \\
            \midrule
            Time (ms) & 5.853 & 1.27 & 7.659 & 16.272 \\
            \bottomrule
        \end{tabular}
    \end{center}
\end{table}
Figure~\ref{visualization} presents a qualitative comparison of the predicted distributions on the Twitter-LDL dataset.
The visualization encompasses two aspects:
1. different scenarios, such as human, animal, etc.
2. the impact of style on emotions.
From the illustration, we could discern that our method has achieved decent prediction results.
In particular, our model can identify well the changes in emotions induced by stylistic information.
Taking the first and second images in Figure~\ref{visualization} as an example, the second one evokes a more melancholic state, and our method cannot solely rely on content perception alone to account for this difference, thereby substantiating the efficacy of incorporating style information.

\section{CONCLUSION}
In this paper, we propose a novel image emotion distribution learning method termed StyleEDL to learn emotional distribution in a style-induced manner.
In StyleEDL, we sought stylistic-aware representations of images based on the hierarchical stylistic information of visual parts. 
In addition, emotional-aware enhanced representations are obtained and further exploited to explore correlations between emotions by the stylistic GCN.
Comprehensive experiments on three well-known datasets demonstrate the superiority of our StyleEDL.

\clearpage
\bibliographystyle{ACM-Reference-Format}
\balance
\bibliography{mm}










\end{document}